%% file: main.tex
\documentclass{article} 
\usepackage{main,times}

\input{math_commands.tex}

\usepackage[utf8]{inputenc} 
\usepackage[T1]{fontenc}    
\usepackage[colorlinks=true, linkcolor=blue, citecolor=blue, urlcolor=blue]{hyperref}
\usepackage{url}
\usepackage{booktabs}       
\usepackage{amsfonts}       
\usepackage{nicefrac}       
\usepackage{microtype}      
\usepackage{xcolor}         
\usepackage{wrapfig}
\usepackage{graphicx}
\usepackage{listings}
\usepackage{wrapfig}
\usepackage{caption}
\usepackage{enumitem}
\usepackage{array}
\usepackage[most]{tcolorbox}

\makeatletter
\renewcommand{\chapterautorefname}{\S\@gobble} 
\renewcommand{\sectionautorefname}{\S\@gobble} 
\renewcommand{\subsectionautorefname}{\S\@gobble} 
\renewcommand{\subsubsectionautorefname}{\S\@gobble} 
\renewcommand{\appendixautorefname}{\S\@gobble} 
\renewcommand{\cite}{\citep}
\makeatother

\title{Efficient Agent Training for Computer Use}

\author{Yanheng He\textsuperscript{1,3}\thanks{Equal contribution.~$^\dagger$Corresponding author.} \quad
Jiahe Jin\textsuperscript{1,3}$^*$ \quad
Pengfei Liu\textsuperscript{1,2,3}$^\dagger$ 
\\
\textsuperscript{1}Shanghai Jiao Tong University \quad
\textsuperscript{2}SII \quad
\textsuperscript{3}GAIR
}

\iclrfinalcopy 
\begin{document}

\maketitle

\begin{abstract}

Scaling up high-quality trajectory data has long been a critical bottleneck for developing human-like computer use agents. We introduce PC Agent-E, an efficient agent training framework that significantly reduces reliance on large-scale human demonstrations. Starting with just \textbf{312} human-annotated computer use trajectories, we further augment them by synthesizing diverse alternative action decisions with Claude 3.7 Sonnet. Trained on these enriched trajectories, our PC Agent-E model achieved a remarkable \textbf{141\%} relative improvement, and even surpassed the Claude 3.7 Sonnet by \textbf{10\%} in relative terms on WindowsAgentArena-V2, an improved benchmark we also released. By integrating robust human computer use skills with automated AI data synthesis capabilities, our method not only brought substantial improvements over training on human trajectories alone, but also significantly surpassed direct distillation from Claude 3.7 Sonnet. Code, data and models are available at \url{https://github.com/GAIR-NLP/PC-Agent-E}.

\end{abstract}

\section{Introduction}
\label{sec: intro}

\begin{wrapfigure}{r}{0.48\linewidth}
  \vspace{-\baselineskip}
  \centering
  \includegraphics[width=\linewidth]{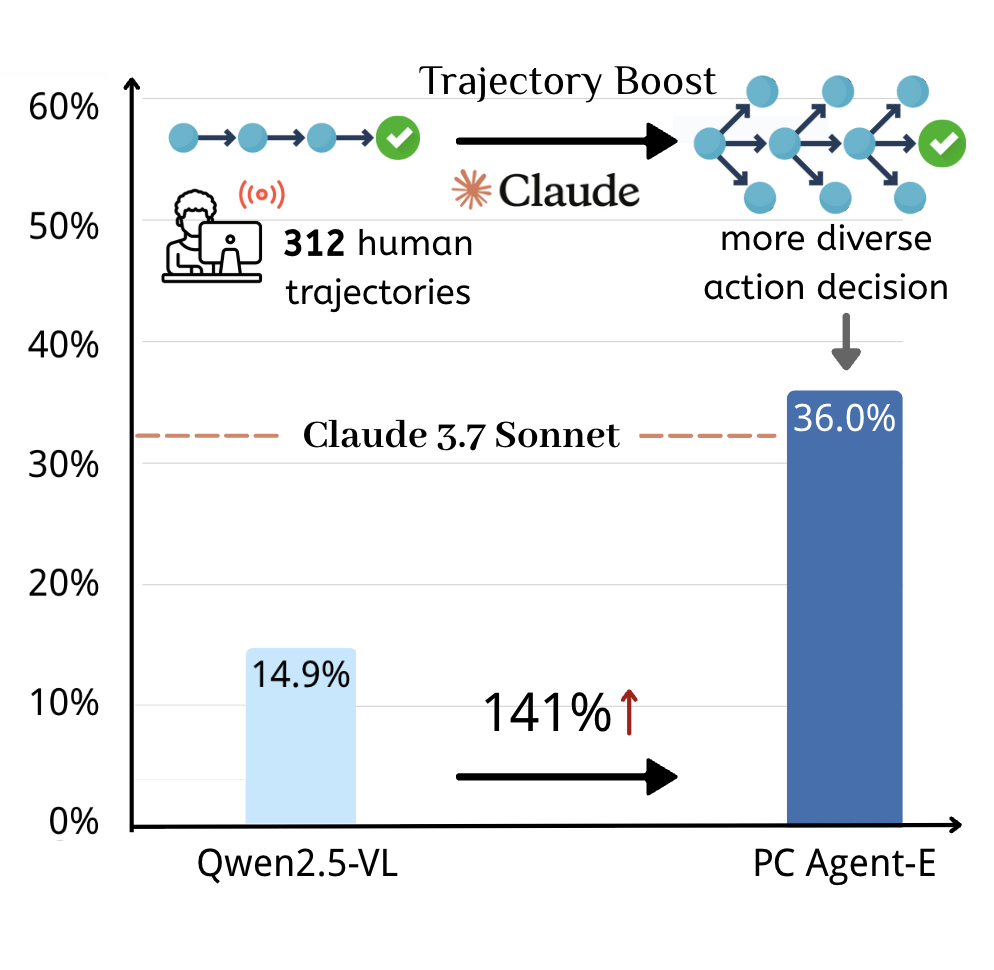}
  \caption{PC Agent-E achieves state-of-the-art open-source performance in Windows computer use with just 312 augmented trajectories.}
  \label{fig:first_image}
\end{wrapfigure}

Developing autonomous agents that can operate computers as humans do~\cite{2024claudecomputeruse,2025openaicua} has long been a landmark pursuit in Artificial Intelligence (AI). Such computer use agents, powered by Vision-Language Models (VLM), perceive screenshots to interact directly with Graphical User Interfaces (GUI) — clicking buttons, navigating menus, and entering text. This allows them to automate a wide range of digital tasks, ranging from routine paperwork and online shopping to complex content creation, promising a significant reduction in manual human workload.

However, current models still fall significantly short of human performance~\cite{xie2024osworldbenchmarkingmultimodalagents,bonatti2024windowsagentarenaevaluating}. This capability gap is even more pronounced within the open-source community, which lacks any solution competitive with leading proprietary systems like Claude 3.7 Sonnet~\cite{2025claude37}. Instilling these advanced computer use capabilities in open-source models remains an unsolved problem. A key factor contributing to these deficiencies is the extreme scarcity of high-quality computer use trajectory data~\cite{ou2024synatraturningindirectknowledge,xu2025agenttrekagenttrajectorysynthesis}.

In this work, we explore efficient agent training for computer use, enabling open-source models to even exceed the performance of proprietary counterparts with minimal human annotation. Inspired by recent findings~\cite{huang2024o1,muennighoff2025s1,ye2025limo} that synthesizing high-quality data using advanced reasoning models like Deepseek-R1~\cite{guo2025deepseek} can efficiently enhance LLM reasoning, we extend the similar idea to the field of computer use agents.

We propose \textbf{PC Agent-E}, an efficient agent training framework that integrates human expertise with AI automation. Starting from a small set of real-world human computer use trajectories, we leverage a frontier agent model to diversify action decisions, yielding richer supervision signals. Training on these augmented trajectories, our agent demonstrates strong computer use capabilities with remarkable data efficiency.

We begin by collecting \textbf{312} human computer use trajectories with PC Tracker~\cite{he2024pcagentsleepai}, a tool for gathering human-computer interaction data, with only two humans annotating one day. These trajectories include task descriptions, screenshots, and human keyboard/mouse actions. We subsequently reconstruct the implicit thought process behind human actions, obtaining comprehensive human trajectories with thoughts. The successful completion of these tasks is inherently assured by human proficiency, obviating the need for additional verification.

While these human trajectories already serve as valuable agent training data, we further augment them through \textbf{Trajectory Boost}, a data synthesis method we developed to enrich each trajectory step with diverse alternative action decisions. The key insight is that computer use tasks can be completed through multiple valid pathways, meaning each step has various reasonable action alternatives supported by rational thought. To capture this diversity, we use a strong agent model to synthesize other possible action decisions. 
Specifically, each human trajectory step provides an \textbf{\textit{environment snapshot}} that encapsulates the essential state information necessary for computer use agents to make decisions. We provide these snapshots to Claude 3.7 Sonnet and sample multiple possible action decisions, thereby significantly enriching and diversifying the trajectory data.

Experimental results demonstrate that with only a small set of human-annotated trajectories, our method can boost the performance of an open-source model to that of frontier models.
Trained with only 312 trajectories augmented by Claude 3.7 Sonnet, our PC Agent-E model achieves an impressive \textbf{141\%} relative improvement over the base model Qwen2.5-VL-72B and even outperforms the teacher model Claude 3.7 Sonnet by 10\% in relative terms on \textbf{WindowsAgentArena-V2}, a benchmark we improved from WindowsAgentArena~\cite{bonatti2024windowsagentarenaevaluating}. Furthermore, PC Agent-E generalizes well to different operating systems on OSWorld~\cite{xie2024osworldbenchmarkingmultimodalagents}. 
Our ablation study further demonstrates that our method for utilizing human demonstrations is not only superior to relying solely on the human trajectories but is also significantly more effective and efficient than directly distilling from the teacher model.

\begin{figure}
    \centering
    \includegraphics[width=1\linewidth]{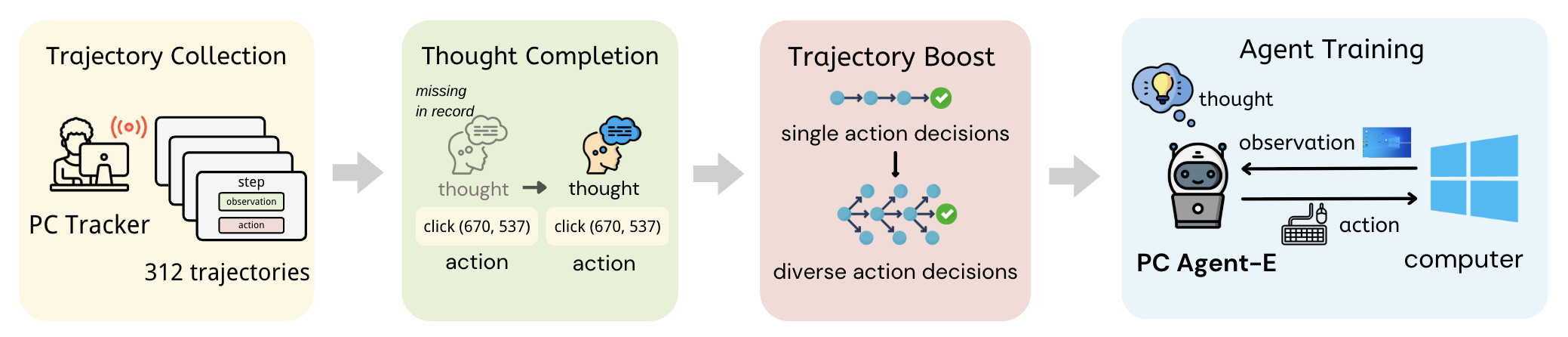}
    \caption{Overview of our framework, consisting of four key components: (1) Trajectory Collection, gathering a small set of human trajectories by recording user actions and state observations at each step; (2) Thought Completion, reconstructing the implicit thought process missing in raw human trajectories; and (3) Trajectory Boost, diversifying action decisions to further augment trajectories (4) Agent Training, developing a strong computer use agent with remarkable data efficiency.}
    \label{fig: method overview}
\end{figure}

In summary, our key contributions are threefold:

\begin{enumerate}
    \item We propose \textbf{Trajectory Boost}, a simple data synthesis method that unlocks remarkable data efficiency for training computer use agents. 
    By augmenting human trajectories with diverse action decisions from a frontier model, our method demonstrates significantly greater effectiveness and efficiency than both using human data alone and direct distillation.
    \item We release \textbf{WindowsAgentArena-V2}, an improved benchmark rectifying evaluation dependence, infeasible hacking, and other limitations in the original WindowsAgentArena benchmark, ensuring more robust and fair evaluations of computer use.
    \item We developed \textbf{PC Agent-E}, an open-source computer use agent that achieves performance comparable to leading proprietary models. Trained with just 312 augmented trajectories, our model successfully outperforms the strong teacher model, Claude 3.7 Sonnet, showing exceptional data efficiency.
\end{enumerate}

\section{Related Work}
\label{gen_inst}

\subsection{Computer Use Agent}


With the advancement of VLMs~\cite{bai2025qwen25vltechnicalreport,deitke2024molmo}, the way computer use agents interact with computers has gradually shifted from relying on textual representations such as accessibility trees~\cite{2024agents,wu2024oscopilot} to directly using screenshots~\cite{uitars2025, xu2025aguvisunifiedpurevision, he2024pcagentsleepai}. Existing computer use agents can be categorized based on how much human prior is built into their design: One is \textit{modular agent workflows}~\cite{2024agents,wu2024oscopilot}, which defines specialized modules and prompts multi-agents to collaborate. The other is \textit{native agent models}~\cite{2024claudecomputeruse,uitars2025,2025openaicua}, which depends on a single model to take action step by step based on its history and current state.

While modular agent workflows can reduce task complexity, their heavy reliance on human priors hinders adaptation to new domains and limits end-to-end optimization~\cite{SaltzerReedClark1984end2end,pan2024swegym}. With the continuous enhancement of model capabilities, native agent models have emerged as the dominant paradigm. This approach offers flexibility, generalizability, and sustainable gains via supervised fine-tuning (SFT)~\cite{xu2025aguvisunifiedpurevision} or reinforcement learning (RL)~\cite{2025openaicua}. Our work explores the efficient agent training methods for native agent models through SFT.

\subsection{Data Synthesis}
As Large Language Models (LLMs) grow ever more powerful, it has become a common practice to use them to synthesize data. Distillation methods~\cite{taori2023alpaca, gunasekar2023textbooksneed, xu2023wizardlmempoweringlargelanguage} leverage state-of-the-art (SOTA) models to generate large-scale data for training weaker models. On the other hand, self-improvement methods enable a model to bootstrap and refine its own training data~\cite{wang2023selfinstructaligninglanguagemodels2}.

In the domain of computer use agents, data synthesis can be broadly categorized into three aspects: (1) large-scale datasets that build foundational GUI understanding, with tasks like screenshot captioning or question–answering~\cite{liu2024harnessingwebpageuistextrich, uitars2025} (2) single-step visual grounding, where mouse click tasks are synthesized from specific locations on the GUI~\cite{gou2025navigatingdigitalworldhumans, wu2024osatlasfoundationactionmodel}
 (3) multi-step trajectory, in which recent research has explored leveraging web tutorials to guide trajectory generation~\cite{ou2024synatraturningindirectknowledge, xu2025agenttrekagenttrajectorysynthesis} or reverse-synthesizes tasks from the agents' own exploration records~\cite{sun2025osgenesisautomatingguiagent, murty2024nnetscape}. 
 Our work differs from prior work by synthesizing high-quality multi-step trajectories based on real-world human demonstrations and emphasizing data efficiency.

\section{Method}

\subsection{Overview}

We propose {PC Agent-E}, an efficient agent training framework for computer use that integrates human expertise with AI automation, as illustrated in Figure~\ref{fig: method overview}. Our method generates high-quality trajectory data by combining authentic human-computer interactions with diverse action decisions, offering advantages in both realism and diversity.

\begin{enumerate}
    \item First, we gathered a small set of 312 task trajectories from human annotators, recording both the screen state observation and the human action at each step, and then filtered the data to remove erroneous steps and trajectories. (\autoref{sec: traj collection})
    \item Subsequently, we reconstructed the latent human thought process before each action decision, based on the corresponding screen state observation and history step context. (\autoref{sec: thought completion})
    \item Then, using human trajectories as \textit{environment snapshots}, we employ Claude 3.7 Sonnet to synthesize diverse alternative action decisions with the Trajectory Boost method. (\autoref{sec:trajboost}) 
    \item Finally, we develop PC Agent-E, our SOTA native agent model for Windows computer use, trained on our augmented trajectories with a simple end-to-end scaffold. (\autoref{sec: agent training})
\end{enumerate}

\subsection{Trajectory Collection}
\label{sec: traj collection}

\begin{wrapfigure}{r}{0.4\linewidth}
  \centering
  \includegraphics[width=\linewidth]{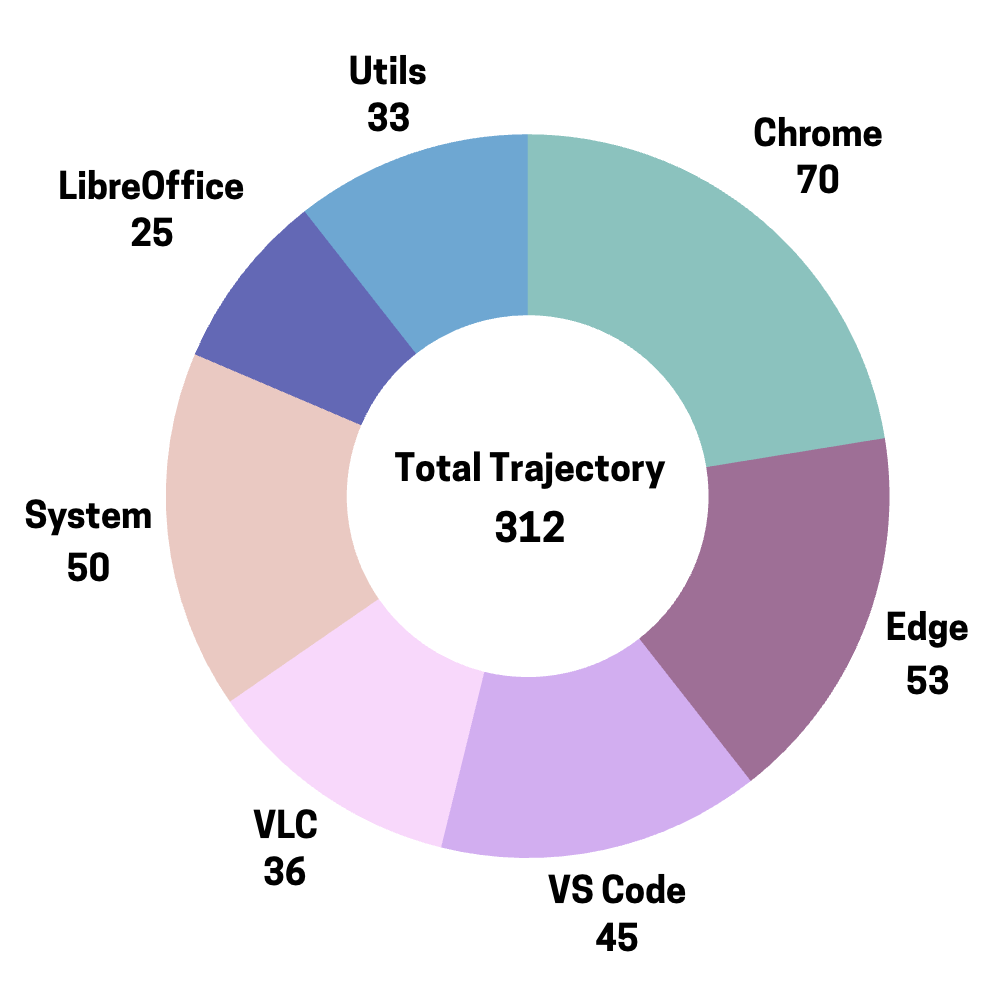}
  \caption{Distribution of the 312 task trajectories across different applications.}
  \label{fig:data distribution}
  \vspace{1.5em}
  \begin{tabular}{lcc}
  \toprule
  \textbf{Length Range} & \textbf{Count} & \textbf{Percentage} \\
  \midrule
  2--5   & 74  & 23.7\% \\
  6--10  & 139 & 44.6\% \\
  11--15 & 73  & 23.4\% \\
  16--30 & 26  & 8.3\%  \\
  \bottomrule
  \end{tabular}
  \captionof{table}{Distribution of the 312 task\newline trajectories across length ranges (steps).}
  \label{tab:traj_length_dist}
\end{wrapfigure}

We collected human computer use trajectories with PC Tracker~\cite{he2024pcagentsleepai}, a tool that records the screen state observation and the human keyboard/mouse action at each step for a given task. The recorded actions are structured in a unified action space $\mathcal{A}$, as shown in Table~\ref{tab:action-space}. For task generation, we first manually compose a small seed set across multiple software applications and then enlarge it with LLMs. The resulting tasks were distributed to human annotators, who completed the tasks on their own Windows computers with PC Tracker recording trajectories automatically. After finishing a task, annotators could either discard unsatisfactory trajectories or modify the task descriptions based on their actual execution, thereby ensuring the correctness and completeness of the collected trajectories. We then applied a set of rule-based filters to remove entire trajectories or individual steps that exhibited errors or other undesirable behaviors. 

We employed a rigorous data decontamination procedure on these collected trajectories. Each task description was compared against the tasks in our main evaluation benchmark (\autoref{sec:waa-v2}) using n-gram overlap and semantic similarity metrics.
Specifically, we retained only trajectories whose task descriptions have an overlap of 0 under 13-gram and less than 0.7 under 3-gram, and a cosine semantic similarity under 0.85 with respect to any test task.

This procedure finally yielded 312 real-world human computer use trajectories, with distribution across applications shown in Figure~\ref{fig:data distribution}. We report the trajectory length distribution in Table~\ref{tab:traj_length_dist}. No explicit instructions on trajectory length were given to annotators; the distribution naturally emerges from the collected tasks and human demonstration behavior. The whole annotation process was completed by two annotators within a single day, with an average of roughly 3 minutes per trajectory. Given humans' proficiency in computer use, the mechanisms for annotators to discard trajectories or revise task descriptions after execution, and our data filtering process, no additional verification was required to ensure trajectory correctness.

\begin{wraptable}{r}{0.5\textwidth}
\centering
\small
\resizebox{\linewidth}{!}{%
\begin{tabular}{@{}lp{4.5cm}@{}}
\toprule
\textbf{Action} & \textbf{Description} \\ \midrule
\textit{click (x, y)} & clicks at coordinates. \\
\textit{right click (x, y)} & right-click at coordinates. \\
\textit{double click (x, y)} & double-click at coordinates. \\
\textit{drag from (x1, y1) to (x2, y2)} & drag the mouse. \\
\textit{scroll (x)} & scrolls the screen with offset x. \\
\textit{press key: enter} & presses the Enter key. \\
\textit{hotkey (ctrl, c)} & performs the Ctrl+C hotkey. \\
\textit{type text: hello} & type text ``hello''. \\
\textit{wait} & pauses for some time. \\
\textit{finish} & the task is finished. \\
\textit{fail} & the task is failed. \\ \bottomrule
\end{tabular}%
}
\caption{Unified action space $\mathcal{A}$.}
\label{tab:action-space}
\end{wraptable}

\subsection{Thought Completion}
\label{sec: thought completion}

We first reconstruct the implicit thought process behind human actions using an iterative approach. Specifically, for each action in the raw trajectory, we provide Claude 3.7 Sonnet with: task description, historical actions with their previously reconstructed thought processes, the current action, and the corresponding screenshot. Based on this information, the model generates the implicit thought process behind action. As shown in Figure~\ref{fig:trajboost}, the recorded raw human trajectory was converted to a human trajectory with thoughts, where the reconstructed thought process is added to each step. See our prompt in Appendix~\ref{appendix: thought completion}.

\subsection{Trajectory Boost}
\label{sec:trajboost}

After thought completion, we obtain comprehensive human trajectories with explicit thought processes. While these trajectories already serve as valuable agent training samples, we further augment them through a simple but effective approach called \textbf{Trajectory Boost}, which synthesizes diverse alternative action decisions for each step of the trajectory.

The motivation behind Trajectory Boost is that computer use tasks inherently allow for multiple valid solution pathways. Consequently, at any given step, several reasonable actions supported by rational thought processes may exist, extending beyond the single solution adopted by human annotators. To capture this inherent diversity, we utilize a frontier computer use agent model, Claude 3.7 Sonnet, to generate single-step alternative action decisions. Its long-horizon planning capabilities, advanced reasoning patterns, and broad knowledge of computer use enable it to generate thought processes and actions that are highly informative and valuable, thereby substantially enhancing the richness and diversity of our trajectory data.

\begin{figure}
    \centering
    \includegraphics[width=1\linewidth]{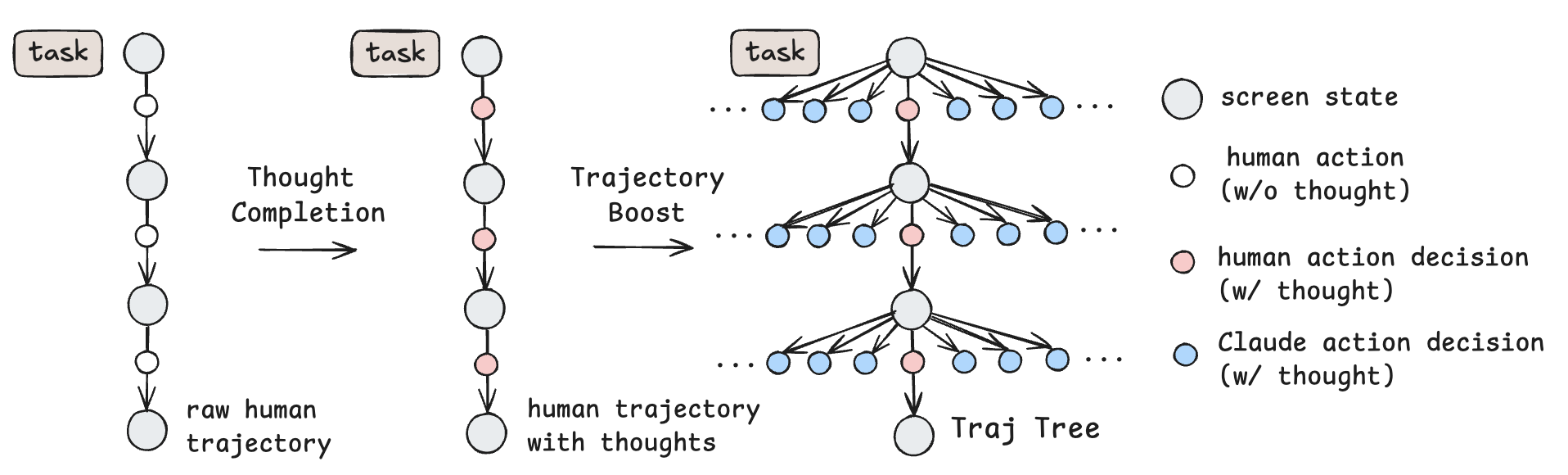}
    \caption{Visualization of our Trajectory Boost method. (Left) Raw human trajectory recorded by PC Tracker. (Center) Human trajectory with reconstructed thoughts after Thought Completion, where the red node indicates human action decisions. (Right) The final \textit{Traj Tree}, where the blue node indicates augmented diverse action decisions synthesized by Claude 3.7 Sonnet. }
    \label{fig:trajboost}
\end{figure}

Specifically, we recognize that each step in a human trajectory captures an \textit{environment snapshot} of the computer, providing the necessary information for both humans and agents to make decisions. For step $k$ on a human trajectory with observation $o_k$, thought process $t_k$, action $a_k$ and task description $T$, the \textit{environment snapshot} is $<T,o_k, h_k>$, where the history context $h_k = (t_1, a_1, t_2, a_2, \dots, t_{k-1}, a_{t-1})$ is constructed with previous human steps. We input this \textit{environment snapshot} to Claude 3.7 Sonnet instantiate in the PC Agent-E scaffold (\autoref{sec: agent training}), and sample multiple single-step action decisions $(t_k'$, $a_k')$ from it. Prompts used are shown in Appendix~\ref{appendix: traj boost}. In practice, we sample 9 action decisions in parallel. This produces a \textbf{\textit{Traj Tree}}, as shown in Figure~\ref{fig:trajboost}, with human trajectory forming the main trunk and the augmented action decisions branching off as leaf nodes. These sampled action decisions from Claude 3.7 Sonnet are not executed in real computer environments, but serve as important augmented data for later agent training.

\subsection{Agent Training}
\label{sec: agent training}

\begin{figure}[h]
    \centering
    \includegraphics[width=1\linewidth]{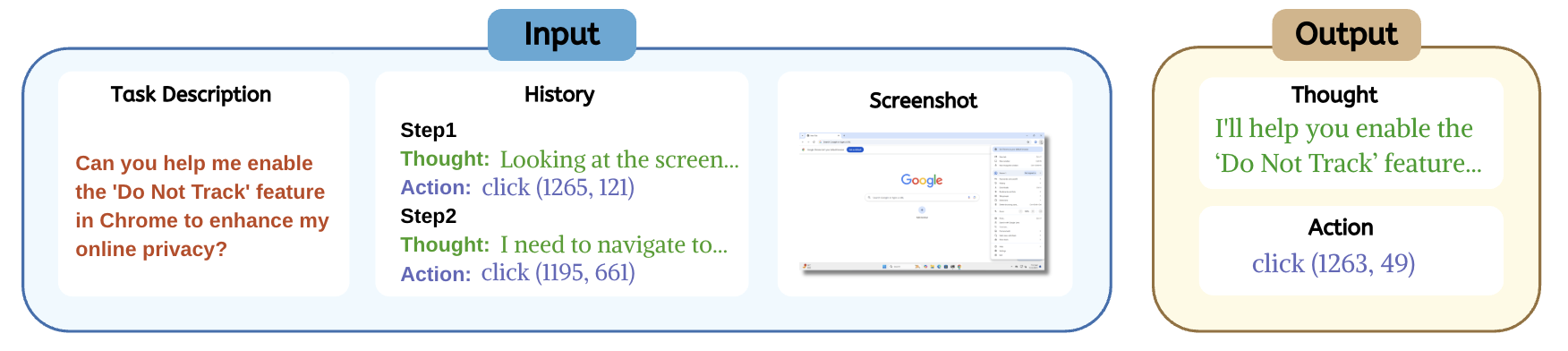}
    \caption{A training example that also demonstrates the inference process of the PC Agent-E scaffold.}
    \label{fig:scaffold}
\end{figure}

PC Agent-E adopts a deliberately simple end-to-end scaffold, as our primary focus is on validating the effectiveness of our agent training methodology rather than optimizing performance through complex workflow design or elaborate prompt engineering. At inference, PC Agent-E takes ⟨screenshot, task description, history⟩ as input and outputs a ⟨thought, action⟩ decision in the ReAct~\cite{yao2023reactsynergizingreasoningacting} paradigm, as shown in Figure~\ref{fig:scaffold}. The action space is the same as $\mathcal{A}$ in Table~\ref{tab:action-space}, and every action is executed via the \texttt{PyAutoGUI} library. The history is a textual log of previous thoughts and actions. 
To maintain simplicity in both training and inference, past screenshots are excluded, although we believe that adding this image history would be beneficial for improving model performance. The prompt used for the scaffold is shown in Appendix~\ref{appendix: scaffold}.

For training, we transform each action node from our \textit{Traj Tree} into an individual training sample. The training sample structure and the inference-time scaffold of the agent share a direct correspondence, as illustrated in Figure~\ref{fig:scaffold}. For both human-demonstrated and model-synthesized action nodes, the history in the training sample includes only prior human actions on the main trunk of the \textit{Traj Tree}. This is consistent with the historical context available to both humans and the model when making the corresponding action decision. We finally obtained 27K training samples from 312 augmented trajectories, each following a consistent input-output structure at inference time.

\section{WindowsAgentArena-V2}
\label{sec:waa-v2}

\begin{figure}[t!]
    \centering
    \includegraphics[width=1\linewidth]{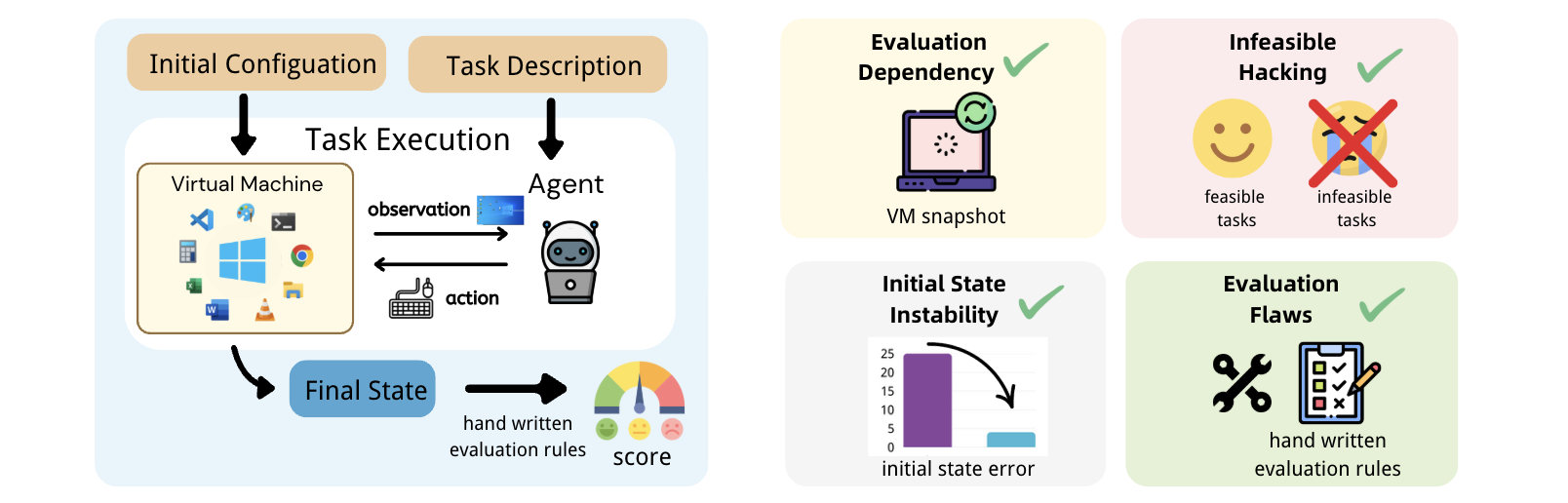}
    \caption{(Left) Overview of the WindowsAgentArena benchmark. (Right) Our main modifications to the updated WindowsAgentArena-V2 benchmark.}
    \label{fig:waa-v2}
\end{figure}

We initially performed our evaluation on WindowsAgentArena~\cite{bonatti2024windowsagentarenaevaluating}, a benchmark designed to assess computer use ability in realistic Windows OS environments through diverse tasks across multiple applications. It provides automatic initial configuration of virtual machine (VM) state and hand-written evaluation rules (see Figure~\ref{fig:waa-v2}). However, we identified several limitations during our assessment. To ensure evaluation reliability, we developed \textbf{WindowsAgentArena-V2}, an updated benchmark comprising 141 tasks across 11 widely-used Windows applications, all derived from the original WindowsAgentArena but with improvements detailed below.

\paragraph{Addressing the evaluation dependency issue.} 
The original benchmark lacked VM state reset between task evaluations, allowing changes from previous tasks to potentially affect subsequent ones. We implemented VM snapshot restoration before each evaluation, ensuring consistent starting states, preventing inter-task interference, and aligning with the i.i.d. (independent and identically distributed) assumption. We also installed some essential software missing from the original VM snapshot but required for proper evaluation.

\paragraph{Preventing infeasible hacking.} 
Current computer use benchmarks such as WindowsAgentArena and OSWorld often include infeasible tasks, which are inherently impossible to complete due to issues such as deprecated system features or user-generated hallucinated commands~\cite{xie2024osworldbenchmarkingmultimodalagents}. The evaluation metric for these tasks is simply considering a task successful if the action \texttt{FAIL} is output at any point during execution. However, we found such evaluation methods particularly easy to hack: an agent can trivially achieve a perfect score on infeasible tasks by always outputting \texttt{FAIL}, without demonstrating any meaningful computer use capabilities. In contrast, completing a feasible task typically requires the agent to execute actions step-by-step to actually fulfill the task objective, posing a significantly different level of difficulty.

We refer to this phenomenon as \textbf{\textit{infeasible hacking}}, a vulnerability confirmed by our subsequent experiments (\autoref{subsec: osworld}), in which a weaker model achieved markedly higher scores on infeasible tasks. Since agents receive identical scores for feasible and infeasible tasks, their coexistence undermines benchmark fairness. Additionally, given that current computer use agents' capabilities are far from optimal, we argue it is presently more valuable to focus on enhancing agents' performance on feasible tasks. Therefore, as a temporary solution in WindowsAgentArena-V2, we removed all infeasible tasks to prevent \textit{infeasible hacking}.

\paragraph{Guaranteeing VM initial state stability.} We found that the state of VM after task initial configuration often exhibited errors like unstable network connections, software launch failures, or system lags. To address this, we designed a validation framework combining rule-based and LLM-based evaluations to verify the initial state, with a re-test mechanism allowing up to three restart attempts for faulty initializations. This approach reduced the initialization failure rate from 10\%–30\% (depending on hardware) to below 5\%.

\paragraph{Fixing evaluation flaws.} 
We discovered that some evaluation functions contained bugs or lacked robustness. For instance, in the task ``\texttt{clearing YouTube history to facilitate finding other histories}'', the evaluation erroneously awarded full scores to agents that deleted the entire browsing history, clearly contradicting user intent. We identified and corrected several evaluation errors and relied on human evaluators for a few complex tasks to improve assessment reliability.

\section{Experiment}
\label{sec: experiment}

In this section, we conduct extensive experiments to evaluate PC Agent-E and validate our Trajectory Boost method. Our experiments are designed to answer the following key questions:

\begin{enumerate}
    \item How does PC Agent-E perform against SOTA methods on computer use tasks?~(\autoref{subsec:main_res})
    \item How does Trajectory Boost's data scaling surpass using human demonstrations alone?~(\autoref{subsec:train_data_scaling})
    \item How does Trajectory Boost differ from and outperform Direct Distillation? (\autoref{subsec:distill_comparison})
    \item How does test-time scaling affect the performance of PC Agent-E?~(\autoref{subsec:test_time_scaling})
    \item How well does PC Agent-E generalize to unseen environments?~(\autoref{subsec: osworld})
\end{enumerate}

\subsection{Setup}
\label{sec: setup}

\paragraph{Benchmarks}

We use WindowsAgentArena-V2 (\autoref{sec:waa-v2}) for the main evaluation, as our training data were collected on the Windows system. We also include results on the original WindowsAgentArena \cite{bonatti2024windowsagentarenaevaluating} in Appendix~\ref{appendixb} for completeness. To test generalization across operating systems, we report results on OSWorld~\cite{xie2024osworldbenchmarkingmultimodalagents}, another computer use benchmark for Linux systems.

\paragraph{Model Baseline}

We compare PC Agent-E with several SOTA models. These include the leading proprietary models Claude 3.7 Sonnet~\cite{2025claude37} and Claude 3.7 Sonnet with extended thinking~\cite{2025claudethinking}, as well as open-source models UI-TARS~\cite{uitars2025}, UI-TARS-1.5~\cite{seedtars2025}, and Qwen2.5-VL-72B~\cite{bai2025qwen25vltechnicalreport}. We also compared with the popular GPT-4o~\cite{openai2024gpt4o} model.

\paragraph{Method Baseline}

We compare our Trajectory Boost method with two alternative training approaches. The first is standard behavior cloning on the 312 human trajectories after thought completion. The second is direct distillation from Claude. We sample 10 end-to-end trajectories from Claude 3.7 Sonnet for each of the 312 tasks. 
The resulting 3,120 trajectories are then used for training with the identical procedure as PC Agent-E, matching the trajectory number of our method for a fair comparison.

\paragraph{Settings}

All experiments and models utilized a screenshot-only observation setting with a uniform screen resolution of $1280 \times 720$. For the UI-TARS model series, we adopted their native framework, which supports image history and code block actions. All other models, including Claude and Qwen, were evaluated using our simple PC Agent-E scaffold. The default maximum number of steps was set to 30, and we also investigated the impact of varying this step limit on model performance.

\paragraph{Training}

We train our PC Agent-E model based on the Qwen2.5-VL-72B~\cite{bai2025qwen25vltechnicalreport} backbone with 27k data mentioned in \autoref{sec: agent training}. Experiments on a smaller model Qwen2.5-VL-7B are also included in Appendix~\ref{appendixc}. We set the image resolution to $1280 \times 720$ and context length to 8,192 tokens. Further training details can be found in Appendix~\ref{appendixa}.

\subsection{Main Results}
\label{subsec:main_res}

As shown in Table~\ref{tab:main}, PC Agent-E achieves a remarkable \textbf{141\%} relative improvement over the base model Qwen2.5-VL-72B on WindowsAgentArena-V2, even surpassing the strong teacher model Claude 3.7 Sonnet by \textbf{10\%} in relative terms (36.0 vs. 32.6), establishing itself as the SOTA open-source model for Windows computer use. Notably, Claude 3.7 Sonnet used to synthesize our training data did not have the thinking mode enabled, but PC Agent-E achieves performance comparable to the stronger Claude 3.7 Sonnet with extended thinking.

\begin{table}[htbp]
\centering
\resizebox{\textwidth}{!}{%
\begin{tabular}{lcccccccc}
\toprule
\textbf{Models} & \textbf{Libreoffice} & \textbf{Chrome} & \textbf{Edge} & \textbf{System} & \textbf{VS Code} & \textbf{VLC} & \textbf{Utils} & \textbf{Total} \\
\midrule
Number of Tasks                         & 42  & 17  & 13 & 24 & 19 & 14 & 12 & 141 \\
\midrule
GPT-4o	& 0.0 &	5.9	& 0.0 & 8.3	& 0.0 & 0.0	& 0.0 & 2.1 \\
Qwen2.5-VL-72B                             & 0.0 & 34.7 & 15.4 & 20.8 & 26.3 &  7.6 & 16.7 & 14.9 \\
UI-TARS-1.5-7B                                & \textbf{7.1} & 34.7 & 23.1 & 45.8 & 21.1 & 7.6 & 16.7 & 21.3 \\
UI-TARS-72B-DPO                                & 0.0 & 40.6 & 38.5 & 58.3 & 36.8 &  7.6 & 25.0 & 26.2 \\
Claude 3.7 Sonnet                  & 2.4 & 46.5 & \textbf{61.5} & 54.2 & 52.6 & 29.0 & 16.7 & 32.6 \\
Claude 3.7 Sonnet (thinking)          & 2.4 & 64.1 & 46.2 & \textbf{66.7} & 52.6 & 21.9 & 25.0 & 35.4 \\
\midrule
\textbf{PC Agent-E (Ours)}                 & 4.8 & \textbf{64.1} & 46.2 & 50.0 & \textbf{57.9} & \textbf{35.7} & 
\textbf{33.3} & \textbf{36.0} \\
\bottomrule
\end{tabular}%
}
\caption{Results of success rate (\%) for different models on WindowsAgentArena-V2.}
\label{tab:main}
\end{table}

\paragraph{Analysis}


To gain deeper insight into the specific capabilities enhanced through our training, we conducted a qualitative analysis by examining 50 trajectories that Qwen2.5-VL-72B failed but PC Agent-E succeeded, as well as trajectories where both models failed. We categorized the failure patterns into three types: (1) \textbf{\textit{Knowledge}}: models may lack specific computer use knowledge. For instance, a model might not know how to enable a particular feature in VLC (a media player software). (2) \textbf{\textit{Planning}}: models may make incorrect planning, such as failing to recognize and recover from previous erroneous actions. (3) \textbf{\textit{Grounding}}: models may execute actions that are inconsistent with their plan, primarily manifested as mouse-clicking errors. We found that our improvements primarily stem from enhanced \textit{planning} capabilities. After training, PC Agent-E produces noticeably longer thought processes and demonstrates improved reasoning capabilities in verification and self-correction. We did not observe significant improvements in \textit{knowledge} or \textit{grounding} capabilities.

\subsection{Data Scaling over Human Demonstrations}
\label{subsec:train_data_scaling}

To validate the effectiveness of our Trajectory Boost method, we investigate the relationship between the scale of synthesized data and model performance. We define data \textbf{scaling factor}, $s$, as the ratio of the total action number used for training to the action number in the original human trajectory. For the model trained exclusively on the human demonstrations, the scaling factor is $s=1$. Our final model, PC Agent-E, was trained using 9 synthesized actions and 1 original human action per step, corresponding to the scaling factor $s=9+1=10$.

\begin{wrapfigure}[19]{r}{0.5\linewidth}
    \vspace{-\baselineskip}
    \centering
    \includegraphics[width=1\linewidth]{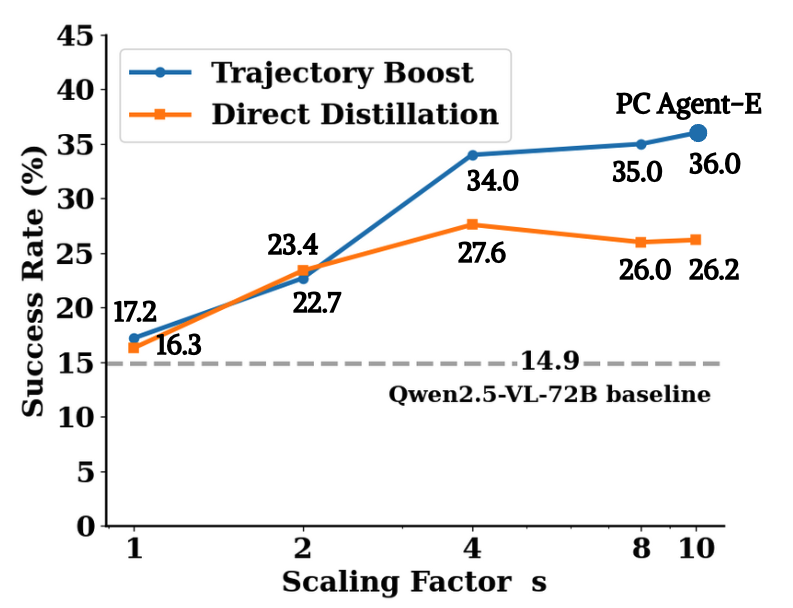}
    \caption{Performance of Trajectory Boost and Direct Distillation method with different data scaling factor $s$ on WindowsAgentArena-V2.}
    \label{fig:scale}
\end{wrapfigure}

As the blue line shown in Figure~\ref{fig:scale}, our results reveal that model performance with the Trajectory Boost method scales significantly with the scaling factor. Compared to training on human trajectories alone, which yields a limited gain (improved from 14.9 to 17.2), PC Agent-E achieves substantially greater performance gains (improved from 14.9 to 36.0). This improvement is primarily driven by the diverse action decisions synthesized from the frontier model with thought processes. This supplements the single human-annotated solution and instills the frontier model's advanced planning capabilities into our agent, thereby yielding performance that far exceeds training on human trajectories alone.

\subsection{Trajectory Boost vs. Direct Distillation}
\label{subsec:distill_comparison}

To demonstrate that our method is more than a simple form of distillation, we compare Trajectory Boost against a direct distillation baseline. For this baseline, we directly sample trajectories from our teacher model, Claude 3.7 Sonnet, end-to-end. We also compare with a variant baseline that combines human demonstrations and direct distillation, with details provided in Appendix~\ref{appendixd:human+distill}.

\paragraph{Superior Performance} As shown in Figure~\ref{fig:scale}, our method significantly outperforms the direct distillation baseline at most of the training data scales (blue line versus orange line). We attribute this to the high quality of our synthesized data. Our method uses human trajectories as a reliable foundation and leverages the frontier model to perform single-step augmentation. This avoids the error accumulation that can occur in end-to-end trajectory distillation.

\paragraph{Exceptional Efficiency}

Another significant advantage of our method is time efficiency. The distillation baseline requires deploying Claude in virtual machines and collecting trajectories through online interaction, which makes it resource-intensive and time-consuming. In contrast, our Trajectory Boost method performs offline data synthesis without interacting with the real environment, enabling natural parallelization. Specifically, to collect an equivalent amount (3120 trajectories) of data, the distillation baseline took about \textbf{900 hours}, while Trajectory Boost required only \textbf{3 hours} under the same hardware conditions — a drastic \textbf{300-fold speedup}.

\subsection{Test Time Scaling}
\label{subsec:test_time_scaling}

\begin{wrapfigure}[13]{r}{0.33\linewidth}
    \vspace{-\baselineskip}
    \vspace{-\baselineskip}
    \centering
    \includegraphics[width=1\linewidth]{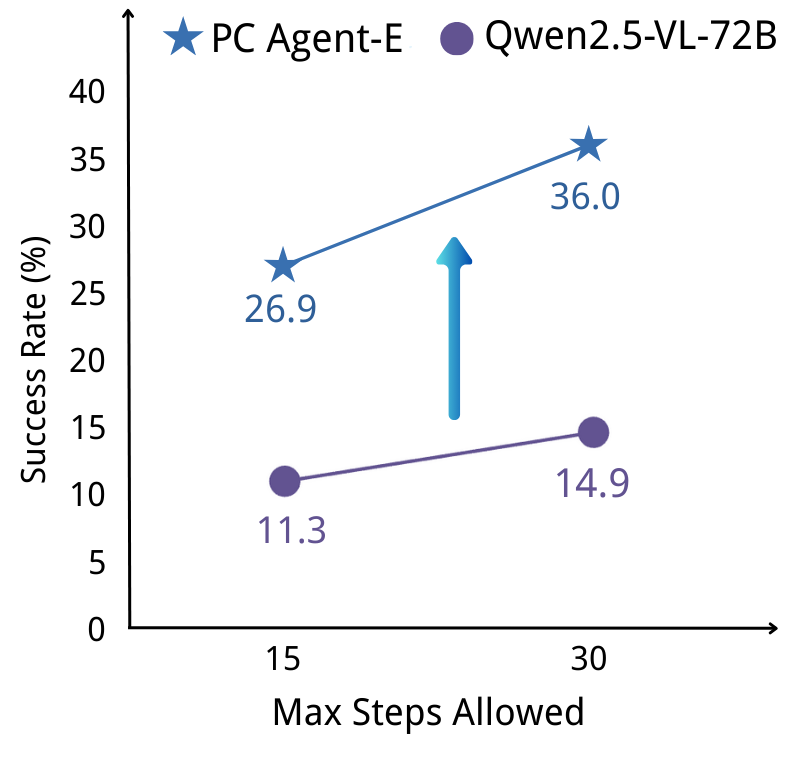}
    \caption{Test time scaling on WindowsAgentArena-V2.}
    \label{fig:tts}
\end{wrapfigure}

We also investigate how the performance of PC Agent-E varies with test time scaling, a topic that has received increasing attention in the research community~\cite{wu2025inferencescalinglawsempirical, snell2024scalingllmtesttimecompute}. We evaluated the model's performance with different numbers of max steps allowed during task completion. As shown in Figure~\ref{fig:tts}, when the step limit increases from 15 to 30, PC Agent-E effectively leverages the additional compute, with the performance gap over the base model widening as more steps are allowed. This suggests that the improved \textit{planning} capabilities acquired through training enable PC Agent-E to benefit from longer interaction horizons. We further extend the evaluation to a 50-step limit in Appendix~\ref{appendix:50step}, where we observe a slight performance decline and analyze potential causes.

\subsection{Cross-Platform Evaluation}
\label{subsec: osworld}

We further evaluated our model on OSWorld to assess cross-platform generalization capabilities. As shown in Table~\ref{tab:osworld}, despite being trained exclusively on Windows data, PC Agent-E achieves a 34\% relative improvement in Linux systems as well. These results validate the generalizability of our method.

\begin{wraptable}[10]{r}{0.5\textwidth}
  \centering
  \vspace{\baselineskip}
  \resizebox{1\linewidth}{!}{
    \begin{tabular}{lccc}
      \toprule
      \textbf{Models} & \textbf{Feasible} & \textbf{Infeasible} & \textbf{Total} \\
      \midrule
      Number of Tasks & 339 & 30 & 369 \\
      \midrule
      Qwen2.5-VL-72B & 4.4 & \textbf{86.7} & 11.1 \\
      \textbf{PC Agent-E (Ours)} & \textbf{10.9} & 63.3 & \textbf{14.9} \\
      \bottomrule
    \end{tabular}%
  }
  \caption{Success Rate (\%) on OSWorld (30-step).}
  \label{tab:osworld}
\end{wraptable}

We also identified an interesting phenomenon in this experiment, which we designate as \textit{infeasible hacking} in~\autoref{sec:waa-v2}: the weaker Qwen2.5-VL-72B model paradoxically achieved markedly better performance on infeasible tasks. This observation suggests that current infeasible task evaluations do not accurately reflect computer use agents' capabilities. Future research may design better criteria for infeasible tasks, such as checking agents' rationale when declaring tasks impossible.

\section{Conclusion}
In this work, we introduce PC Agent-E, an efficient training framework for computer-use agents. Our framework leverages Trajectory Boost, a data synthesis method that augments human demonstrations with diverse action alternatives generated by a frontier model. Trained on as few as 312 augmented trajectories, PC Agent-E achieves a 141\% relative improvement over the base model, even outperforms the strong teacher model Claude 3.7 Sonnet. Ablation studies further demonstrate that this approach is not only significantly superior to using human trajectories alone, but also more efficient and effective than direct distillation from the teacher model. These findings suggest that complex computer-use capabilities can be effectively elicited from a remarkably small set of high-quality trajectories.

\section*{Acknowledgments}

We would like to express our sincere gratitude to Shijie Xia for his meticulous review and constructive suggestions, which significantly improved the quality of this paper. This project is supported by SJTU SEIEE - ByteDance Large Language Model Joint Laboratory, SII.



\bibliography{main}
\bibliographystyle{main}

\newpage
\appendix

\section{Training Details}
\label{appendixa}

The PC Agent-E model is fine-tuned on a dataset of 27k samples for 2 epochs using 32 NVIDIA GPUs over approximately 5 hours, with Qwen2.5-VL-72B as the base model. We set the context length to 8,192 tokens, using a batch size of 128 and a learning rate of 2e-6. The training process employs cosine annealing for learning rate scheduling, with a warm-up ratio of 0.05. The visual tower is kept frozen throughout the training process.

\section{Evaluation On Original WindowsAgentArena Benchmark}
\label{appendixb}

We also evaluate PC Agent-E's performance on the original WindowsAgentArena benchmark. As shown in Table~\ref{tab:original_waa}, our model greatly surpasses the previous SOTA method, NAVI, across all task categories. NAVI is a complex agent framework that integrates GPT-4V~\cite{gpt4} with a specialized tool called Omniparser~\cite{lu2024omniparser}.

\begin{table}[htbp]
\centering
\resizebox{\textwidth}{!}{%
\begin{tabular}{lccccccc}
\toprule
\textbf{Method} & \textbf{Office} & \textbf{Web} & \textbf{System} & \textbf{Coding} & \textbf{Media \& Video} & \textbf{Utils} & \textbf{Overall} \\
\midrule
NAVI~\cite{bonatti2024windowsagentarenaevaluating} & 0.0  & 27.3 & 33.3 & 27.3 & 30.3 & 8.3  & 19.5 \\
\textbf{PC Agent-E (Ours)} & \textbf{2.3}  & \textbf{33.1} & \textbf{70.6} & \textbf{37.5} & \textbf{33.3} & \textbf{25.0} & \textbf{27.9} \\
\bottomrule
\end{tabular}%
}
\caption{Success Rate (\%) on the original WindowsAgentArena benchmark.}
\label{tab:original_waa}
\end{table}

\section{Experiment on Smaller Model}
\label{appendixc}

\begin{wraptable}[8]{r}{0.32\textwidth}
  \vspace{-\baselineskip}
  \centering
  \begin{tabular}{l c}
    \toprule
    \textbf{Method} & \textbf{Overall} \\
    \midrule
    Qwen2.5-VL-7B   & 5.0 \\
    \textbf{PC Agent-E 7B}   & \textbf{6.4} \\
    \bottomrule
  \end{tabular}
  \caption{Success Rate (\%) on WindowsAgentArena-V2.}
  \label{tab:smaller_model}
\end{wraptable}

To test our method on smaller models, we also trained Qwen2.5-VL-7B~\cite{bai2025qwen25vltechnicalreport} on the same 27K dataset for PC Agent-E, resulting in the \textbf{PC Agent-E 7B}. As shown in Table~\ref{tab:smaller_model}, our method yields improvements on the 7B model as well, although the gains are not as significant as those observed with the 72B model. This is because our method mainly enhances the model's \textit{planning} ability, as discussed in~\autoref{subsec:main_res}, but the small model's deficiencies in \textit{knowledge} and \textit{grounding} limit the overall performance gain.

\section{Human Data + Direct Distillation Baseline}
\label{appendixd:human+distill}

To further validate the effectiveness of Trajectory Boost, we compare against a baseline that combines human demonstrations with direct distillation. Specifically, the base model is trained using 312 human trajectories together with 9$\times$312 trajectories sampled end-to-end from Claude 3.7 Sonnet. As shown in Table~\ref{tab:human+distillation}, this combined baseline achieves a score of 26.9, comparable to direct distillation alone (26.2), yet significantly lower than PC Agent-E (36.0).

\begin{table}[h]
\centering
\resizebox{\linewidth}{!}{
\begin{tabular}{lccccccc|c}
\toprule
 & \textbf{LibreOffice} & \textbf{Chrome} & \textbf{Edge} & \textbf{System} & \textbf{VS Code} & \textbf{VLC} & \textbf{Utils} & \textbf{Total} \\
\midrule
Number of Tasks & 42 & 17 & 13 & 24 & 19 & 14 & 12 & 141 \\
\midrule
Direct Distillation & 2.4 & 28.8 & 30.7 & 50.0 & 57.9 & 7.6 & 25.0 & 26.2 \\
Human Data + Direct Distillation & 2.4 & 22.9 & 30.7 & 54.2 & 57.9 & 21.6 & 16.6 & 26.9 \\
\textbf{PC Agent-E (Ours)} & 4.8 & 64.1 & 46.2 & 50.0 & 57.9 & 35.7 & 33.3 & \textbf{36.0} \\
\bottomrule
\end{tabular}
}
\caption{Comparison with Human Data + Direct Distillation baseline on WindowsAgentArena-V2.}
\label{tab:human+distillation}
\end{table}


\newpage
\section{Extended Test-Time Scaling: 50-Step Evaluation}
\label{appendix:50step}

In addition to the 15-step and 30-step evaluations presented in Figure~\ref{fig:tts}, we conduct an extended evaluation with a 50-step limit to further investigate the test-time scaling behavior of PC Agent-E. As shown in Table~\ref{tab:scaling}, both PC Agent-E and the base model Qwen2.5-VL-72B exhibit performance drops when the step limit is extended to 50.

\begin{table}[h]
\centering
\begin{tabular}{lccc}
\toprule
\textbf{Model} & \textbf{15-step} & \textbf{30-step} & \textbf{50-step} \\
\midrule
Qwen2.5-VL-72B & 11.3 & 14.9 & 11.9 \\
PC Agent-E & 26.9 & 36.0 & 31.4 \\
\bottomrule
\end{tabular}
\caption{Success Rate (\%) on WindowsAgentArena-V2 under varying step limits. Performance improves from 15 to 30 steps but degrades at 50 steps for both models.}
\label{tab:scaling}
\end{table}

\paragraph{Analysis.}
We identify that the performance degradation at 50 steps is primarily caused by the models' inability to correctly recognize task completion. Without a robust stopping mechanism, the additional steps allow the agent to continue acting after a task has already been successfully completed, often leading to actions that undo or disrupt the correct result. We attribute this to the training data distribution (see Table~\ref{tab:traj_length_dist} in~\autoref{sec: traj collection}), where over 90\% of trajectories contain fewer than 15 steps, leaving the model with limited exposure to long-horizon execution and insufficient supervision for task termination.

\paragraph{Scaling within and beyond the training distribution.}
Combining the results in Table~\ref{tab:scaling}, we observe two distinct regimes. Within a range reasonably close to the training distribution (15 $\rightarrow$ 30 steps), our method demonstrates effective test-time scaling, with performance improving as more steps are allowed. However, when the step limit is pushed significantly beyond the training distribution (50 steps), performance degrades due to the lack of termination awareness. This suggests that the current training paradigm supports \emph{near-distribution} test-time scaling but does not yet generalize to substantially longer horizons.

\paragraph{Future directions.}
We hypothesize that addressing this limitation requires either (1)~incorporating longer trajectories into the training data to extend the model's effective operating range, or (2)~introducing targeted supervision for task completion detection, such as an explicit termination prediction objective. We leave the exploration of these directions to future work.


\newpage
\section{Prompts}
\label{appendix: prompts}
\subsection{Thought Completion}
\label{appendix: thought completion}

\lstset{
    basicstyle=\ttfamily\small,
    columns=flexible,
    breaklines=true,
    breakindent=0pt,
    literate={`}{{\textasciigrave}}1
             {√}{{\checkmark}}1
             {@}{{\texttimes}}1
}

\setlength{\parindent}{0pt}

\spaceskip=0.3em

\begin{table}[h!]
    \caption{Thought Completion Prompt}
    \label{thought completion}
    \renewcommand{\arraystretch}{1}
    \begin{tabular}{>{\setlength{\parskip}{0.18em}}p{0.95\textwidth}}
    \toprule
    \textbf{Prompt for Thought Completion} \\
    \midrule
    \begin{lstlisting}
    You are a helpful computer use agent designed to complete tasks on a computer. Your goal is to recreate your thought process behind a specific action.

    You will be provided with:
    
    1. The task you are attempting to complete.
    2. A history of the steps you have already performed (up to 50, if any; none if it was the first action).
    3. The specific action you chose to take.
    4. The name of the element you clicked (if you clicked on an element). It might be too general or vague, you have to decied what to click based on the screenshot.
    5. A screenshot of the computer screen at the moment you decided to take the action.
    6. The red marks on the screenshot indicate the position of the click or drag action.
    
    To formulate your thought process, consider:
    
    1. What do you observe on the screen? Consider your task and previous action when you analyzing current screenshot.
    2. Evaluate your previous action (if applicable):
       - Did it achieve the intended effect? If not, identify possible reasons (e.g., misclick, inactive element).
          Some typical examples for ineffective action:
           - misclick in an empty space
           - ineffective opening some elements without double click
           - ineffective type text/ press key because of inactivated input box
       - Did the result align with your previous plan, or did something unexpected happen?
          Some typical examples for ineffective action:
             - misclick in a wrong element
             - forget to clear existing text in input bar
    3. Based on your action history, assess your progress toward completing the overall task.
    4. Consider if you're exploring how to finish the task because of failed attempts in history steps.

    Present your thought process as a clear, natural first-person narrative that explains your reasoning at that moment.

    Important requirements:
    1. **DO NOT** mention the red marks in your response. These marks were **added after the fact** 
    to indicate the position of your click or drag actions, and they were not on the screen when you made the decision. **DO NOT** mention "red box", "red square", "red circle", or "red arrow" in your response.
    
    \end{lstlisting}
    \end{tabular}
    \end{table}
    
    \clearpage
    \vspace*{-\topskip}
    \vspace{0pt}
    \begin{table}[!t]
    \setlength{\aboverulesep}{0pt}
    \setlength{\belowrulesep}{0pt}
    \setlength{\extrarowheight}{0pt}
    \begin{tabular}{>{\setlength{\parskip}{0.2em}}p{0.95\textwidth}}
    \begin{lstlisting}[basicstyle=\ttfamily\small]
    
    2. Write as if you are thinking in real-time before taking the action. Do not include post-action evaluation or hindsight.

    The task you are attempting to complete: {task_description}
    Your performing history: {history_str}
    The specific action you chose to perform: {action}
    \end{lstlisting} \\
    \bottomrule
    \end{tabular}
    \end{table}

\subsection{Trajectory Boost}
\label{appendix: traj boost}

\lstset{
    basicstyle=\ttfamily\small,
    columns=flexible,
    breaklines=true,
    breakindent=0pt,
    literate={`}{{\textasciigrave}}1
             {√}{{\checkmark}}1
             {@}{{\texttimes}}1
}

\setlength{\parindent}{0pt}

\spaceskip=0.3em

\begin{table}[h!]
    \caption{Trajectory Boost Prompt}
    \label{traj boost}
    \renewcommand{\arraystretch}{1}
    \begin{tabular}{>{\setlength{\parskip}{0.18em}}p{0.95\textwidth}}
    \toprule
    \textbf{Prompt for Trajectory Boost} \\
    \midrule
    \begin{lstlisting}
    You are a helpful assistant who can help users complete computer tasks, with **full permission** to make any operations on the user's computer. The operating system is windows.
    Based on the provided current state, you need to suggest the next action to complete the task. Do not try to complete the entire task in one step. Break it down into smaller steps, and at each step you will get a new state to interact with.
    
    IMPORTANT: You must strictly adhere to the following rules:
    
    1. Choose ONLY ONE action from the list below for each response, DO NOT perform more than one action per step.
    2. Follow the exact syntax format for the selected action, DO NOT create or use any actions other than those listed.
    3. Once the task is completed, output action finish.
    
    Valid actions:
    1. click (x, y)
       click the element at the position (x, y) on the screen
    2. right click (x, y)
       right click the element at the position (x, y) on the screen
    3. double click (x, y)
       double click the element at the position (x, y) on the screen
    4. drag from (x1, y1) to (x2, y2)
       drag the element from position (x1, y1) to (x2, y2).
    5. scroll (x)
       scroll the screen vertically with pixel offset x. Positive values of x: scroll up, negative values of x: scroll down.
    6. press key: key_content
       press the key key_content on the keyboard.
    7. hotkey (key1, key2)
       press the hotkey composed of key1 and key2.
    8. hotkey (key1, key2, key3)
       press the hotkey composed of key1, key2, and key3.
    \end{lstlisting}
    \end{tabular}
    \end{table}
    
    \clearpage
    \vspace*{-\topskip}
    \vspace{0pt}
    \begin{table}[!t]
    \setlength{\aboverulesep}{0pt}
    \setlength{\belowrulesep}{0pt}
    \setlength{\extrarowheight}{0pt}
    \begin{tabular}{>{\setlength{\parskip}{0.2em}}p{0.95\textwidth}}
    \begin{lstlisting}[basicstyle=\ttfamily\small] 
    9. type text: text_content
       type content text_content on the keyboard. 
       Note that before typing text, you need to ensure the text box or input field is active/focused first. If the text box is not yet activated, you should first click on it to activate it, and then use type text in a separate step.
    10. wait
        wait for some time, usually for the system to respond, screen to refresh, advertisement to finish.
    11. finish
        indicating that the task has been completed.
    12. fail
        indicating that the task has failed, of this task is infeasible because not enough information is provided.  
    
    Before deciding your next action, you should think carefully about the current state of the screen and your history steps. Contain the following points in your thought process:
    
    1. What do you observe on the screen? Consider your task and previous action when you analyzing current screenshot.
    2. What's your previous plan and action (if applicable)? Evaluate your previous plan and action in three conditions:
       1. It didn't make any effect. You should dentify possible reasons (e.g., misclick, inactive element) and adjust your plan in this step.
          Some typical examples for ineffective action:
           - misclick in an empty space
           - ineffective opening some elements without double click
           - ineffective type text/ press key because of inactivated input box
       2. It made some effect, but the result does not align with previous plan. You should dentify possible reasons (e.g., misclick, inactive element) and correct it in this step.
          Some typical examples for ineffective action:
             - misclick in a wrong element
             - forget to clear existing text in input bar
       3. It made some effect, and it successfully align with previous plan. You should progress to the next step based on the current state.
    3. Based on your action history, assess your progress toward completing the overall task.
    4. Exploring new ways to finish the task if there are already failed attempts in history steps. **DO NOT repeat** the history actions.
    
    Response Format: Your thought process\n\nAction: The specific action you choose to take.

    The task you are attempting to complete: {task_description}
    Your performing history: {history_str}
    Given the screenshot as below. What's the next step that you will do to help with the task?
  
    \end{lstlisting} \\
    \bottomrule
    \end{tabular}
    \end{table}

\newpage
\subsection{PC Agent-E scaffold}
\label{appendix: scaffold}

\lstset{
    basicstyle=\ttfamily\small,
    columns=flexible,
    breaklines=true,
    breakindent=0pt,
    literate={`}{{\textasciigrave}}1
             {√}{{\checkmark}}1
             {@}{{\texttimes}}1
}

\setlength{\parindent}{0pt}

\spaceskip=0.3em

\begin{table}[h!]
    \caption{PC Agent-E scaffold Prompt}
    \label{scaffold prompt}
    \renewcommand{\arraystretch}{1}
    \begin{tabular}{>{\setlength{\parskip}{0.18em}}p{0.95\textwidth}}
    \toprule
    \textbf{Prompt for PC Agent-E scaffold} \\
    \midrule
    \begin{lstlisting}
    You are a helpful assistant who can help users complete computer tasks, with **full permission** to make any operations on the user's computer. 
    Based on the provided current state, you need to suggest the next action to complete the task. Do not try to complete the entire task in one step. Break it down into smaller steps, and at each step you will get a new state to interact with.
    IMPORTANT: You must strictly adhere to the following rules:
    1. Choose ONLY ONE action from the list below for each response, DO NOT perform more than one action per step.
    2. Follow the exact syntax format for the selected action, DO NOT create or use any actions other than those listed.
    3. Once the task is completed, output action finish.
    
    Valid actions:
    1. click (x, y)
    click the element at the position (x, y) on the screen
    
    2. right click (x, y)
    right click the element at the position (x, y) on the screen
    
    3. double click (x, y)
    double click the element at the position (x, y) on the screen
    
    4. drag from (x1, y1) to (x2, y2)
    drag the element from position (x1, y1) to (x2, y2).
    
    5. scroll (x)
    scroll the screen vertically with pixel offset x. Positive values of x: scroll up, negative values of x: scroll down.
    
    6. press key: key_content
    press the key key_content on the keyboard.
    
    7. hotkey (key1, key2)
    press the hotkey composed of key1 and key2.
    
    8. hotkey (key1, key2, key3)
    press the hotkey composed of key1, key2, and key3.
    
    9. type text: text_content
    type content text_content on the keyboard.
    
    10. wait
    wait for some time, usually for the system to respond, screen to refresh, advertisement to finish.
    
    11. finish
    indicating that the task has been completed.
    
    12. fail
    indicating that the task has failed, of this task is infeasible because not enough information is provided.

    \end{lstlisting}
    \end{tabular}
    \end{table}
    
    \clearpage
    \vspace*{-\topskip}
    \vspace{0pt}
    \begin{table}[!t]
    \setlength{\aboverulesep}{0pt}
    \setlength{\belowrulesep}{0pt}
    \setlength{\extrarowheight}{0pt}
    \begin{tabular}{>{\setlength{\parskip}{0.2em}}p{0.95\textwidth}}
    \begin{lstlisting}[basicstyle=\ttfamily\small]
    
    Response Format: {Your thought process}
    Action: {The specific action you choose to take}

    Your task is: {task_description}
    History of the previous actions and thoughts you have done to reach the current screen: {history_str}
    --------------------------------------------
    Given the screenshot, what's the next step you will do to help with the task?
    \end{lstlisting} \\
    \bottomrule
    \end{tabular}
    \end{table}


\newtcbox{\buttonbox}{
    nobeforeafter,
    tcbox raise base,
    boxrule=0.4pt,
    top=2pt,
    bottom=2pt,
    right=4pt,
    left=4pt,
    arc=3pt,
    boxsep=0pt,
    colback=white,
    colframe=black}

\newpage
\section{PC Tracker User Manual}
\label{appendix: manual}

\subsection*{1. Introduction}
PC Tracker is a lightweight infrastructure for efficiently collecting large-scale human-computer interaction trajectories. The program runs seamlessly in the background, automatically capturing screenshots and keyboard \& mouse activities. 

\subsection*{2. Installation}
\begin{itemize}[itemsep=0pt]
    \item Ensure your operating system is Windows.
    \item Extract our software package to a location with sufficient disk space (recommended to have more than 3GB of available space for storing recorded data).
\end{itemize}

\subsection*{3. Quick Start}
\begin{itemize}[itemsep=0pt]
    \item (Optional) Set screen resolution to 16:9.
    \item Open the extracted folder and launch \texttt{main.exe}.
\end{itemize}

\subsection*{4. Instructions}
After launching the tracker, you can choose between \textbf{Task Oriented Mode} or \textbf{Non-Task Oriented Mode} for recording.

\subsubsection*{Task Oriented Mode}
This mode is divided into two sub-modes: \textbf{Given Task} and \textbf{Free Task}.

\paragraph{Given Task}  
In this mode, you will be assigned an uncompleted task each time.
\begin{itemize}[itemsep=0pt]
    \item \textbf{Next Task}: Click \buttonbox{\textsf{Next Task}} to get the next task.
    \item \textbf{Previous Task}: Click \buttonbox{\textsf{Previous Task}} to return to the previous task.
    \item \textbf{Bad Task Feedback}: If you think the current task is difficult to complete, click \buttonbox{\textsf{Bad Task}} to discard it permanently. Alternatively, you can start the task and modify its description after completion based on your actual execution.
    \item \textbf{Start Recording}: Click \buttonbox{\textsf{Start}}, and the tracker window will automatically minimize while recording begins.
    \item \textbf{End Task}: After completing the task, click \buttonbox{\textsf{Finish}} to save the record. Or if the task execution fails or you don't want to record it, click \buttonbox{\textsf{Fail}}.
    \item \textbf{Modify Task Description}: After finishing the task, you can modify the task description based on your actual execution.
\end{itemize}

\paragraph{Free Task}  
In this mode, you can freely use the computer and summarize the task description and difficulty yourself.
\begin{itemize}[itemsep=0pt]
    \item \textbf{Start Recording}: Click \buttonbox{\textsf{Start}}, and the tracker window will automatically minimize while recording begins.
    \item \textbf{Save and Summarize This Record}: 
    Fill in the task description, select difficulty (easy/medium/hard), and click \buttonbox{\textsf{Save}} to save the record.
    \item \textbf{Discard This Record}: Click \buttonbox{\textsf{Discard}} to discard the record.
\end{itemize}

\subsubsection*{Non-Task Oriented Mode}

In this mode, you can freely use the computer, with similar methods to start and stop recording as described above.

\subsection*{5. Usage Notes}
\begin{itemize}[itemsep=0pt]
    \item \textbf{Does not currently support using extended screens.}
    \item \textbf{Does not currently support using Chinese input methods.}
    \item \textbf{Does not currently support using touchpads.}
    \item \textbf{The tracker window is fixed in fullscreen.} To support the filtering of tracker-related actions (such as clicking the Start button) in post-processing, the tracker window is fixed in fullscreen. You can reopen the tracker window by clicking to view the task description, then minimize it again, but please do not drag it to display in a non-fullscreen state.

\end{itemize}

\subsection*{6. Data Privacy}
\begin{itemize}[itemsep=0pt]
    \item After starting recording, your screenshots and keyboard \& mouse operations will be automatically recorded. PC Tracker does not record any information from unopened software. If you believe the recording may infringe on your privacy, you can choose to discard the record.
    \item Collected data is saved in the \texttt{./events} folder (hidden by default). Each trajectory includes a Markdown file for easy visualization.
\end{itemize}

\subsection*{7. FAQ}
\paragraph{Does the tracker have networking capabilities?}  
PC Tracker is completely local, does not support networking, and will not upload your data.

\paragraph{What if my computer screen resolution is not 16:9?}  
If your screen resolution is not 16:9, it will affect the subsequent unified processing of data. We recommend adjusting your screen resolution to 16:9.

\paragraph{How much space does the collected data occupy?}  
The specific data size varies. Generally, even with intensive recording operations for 1 hour, it will not generate more than 1GB of data.

\newpage
\section{The Use of LLMs}
\label{appendix: use of llms}

We used LLMs to improve the grammar, clarity, and overall readability of this paper. All research ideas, content, and scientific contributions were developed and written by the human authors. All suggestions from LLMs were reviewed and edited by the authors, who retain full responsibility for the final content of this paper.

\end{document}

%% file: math_commands.tex
\usepackage{amsmath,amsfonts,bm}

\def\eqref#1{equation~\ref{#1}}

\def\1{\bm{1}}

\DeclareMathAlphabet{\mathsfit}{\encodingdefault}{\sfdefault}{m}{sl}
\SetMathAlphabet{\mathsfit}{bold}{\encodingdefault}{\sfdefault}{bx}{n}

